\documentclass{llncs}

\usepackage{hyperref}
\hypersetup{
    colorlinks=true,
    linkcolor=blue,
    filecolor=magenta,      
    urlcolor=cyan,
}

\usepackage{graphicx}
\usepackage{color}
\usepackage[algo2e,ruled,vlined]{algorithm2e}
\usepackage{amsmath}

\usepackage{amssymb}
\usepackage{subfig}
\usepackage{booktabs}
\usepackage{verbatim}

\DeclareMathOperator*{\argmin}{arg\,min}
\newcommand{\RR}{\mathbb{R}}
\newcommand{\beq}{\begin{equation}}
\newcommand{\eeq}{\end{equation}}
\newcommand{\tr}[1]{{#1}^\top}
\renewcommand{\vec}[1]{\mathbf{#1}}

\newcommand{\tx}[1]{\textrm{#1}}

\newcommand{\yy}{\vec{y}}
\newcommand{\xx}{\vec{x}}
\newcommand{\uu}{\vec{u}}
\newcommand{\zz}{\vec{z}}

\newcommand{\LL}{\mathcal{L}}
\newcommand{\vone}{\vec{1}}
\newcommand{\vnu}{\boldsymbol{\nu}}
\newcommand{\cc}{s}
\newcommand{\qq}{\vec{q}}

\usepackage{tabularx}

\newcolumntype{L}[1]{>{\raggedright\let\newline\\\arraybackslash\hspace{0pt}}m{#1}}
\newcolumntype{C}[1]{>{\centering\let\newline\\\arraybackslash\hspace{0pt}}m{#1}}
\newcolumntype{R}[1]{>{\raggedleft\let\newline\\\arraybackslash\hspace{0pt}}m{#1}}

\usepackage{enumitem}
\setlist[itemize]{noitemsep, topsep=0pt}

\begin{document}

\title{Unbiased Shape Compactness for Segmentation}

\author{Jose Dolz\inst{1}, Ismail Ben Ayed \inst{1}, Christian Desrosiers \inst{1}}

\institute{Laboratory for Imagery, Vision and Artificial Intelligence\\
\'{E}cole de Technologie Sup\'{e}rieure, Montreal, Canada}



\maketitle

\begin{abstract}
We propose to constrain segmentation functionals with a dimensionless, unbiased and position-independent shape compactness prior, which we solve efficiently with an alternating direction method of multipliers (ADMM). Involving a squared sum of pairwise potentials, our prior results in a challenging high-order optimization problem, which involves dense (fully connected) graphs. We split the problem into a sequence of easier sub-problems, each performed efficiently at each iteration: (i) a sparse-matrix inversion based on Woodbury identity, (ii) a closed-form solution of a cubic equation and (iii) a graph-cut update of a sub-modular pairwise sub-problem with a sparse graph.  
We deploy our prior in an energy minimization, in conjunction with a supervised classifier term based on CNNs and standard regularization constraints. We demonstrate the usefulness of our energy in several medical applications. In particular, we report comprehensive evaluations of our fully automated algorithm over $40$ subjects, showing a competitive performance for the challenging task of abdominal aorta segmentation in MRI.
\end{abstract}

\section{Introduction}

\begin{figure}[ht!]
     \begin{center}
     \mbox{
        \includegraphics[width=0.195\linewidth,height=0.195\linewidth]{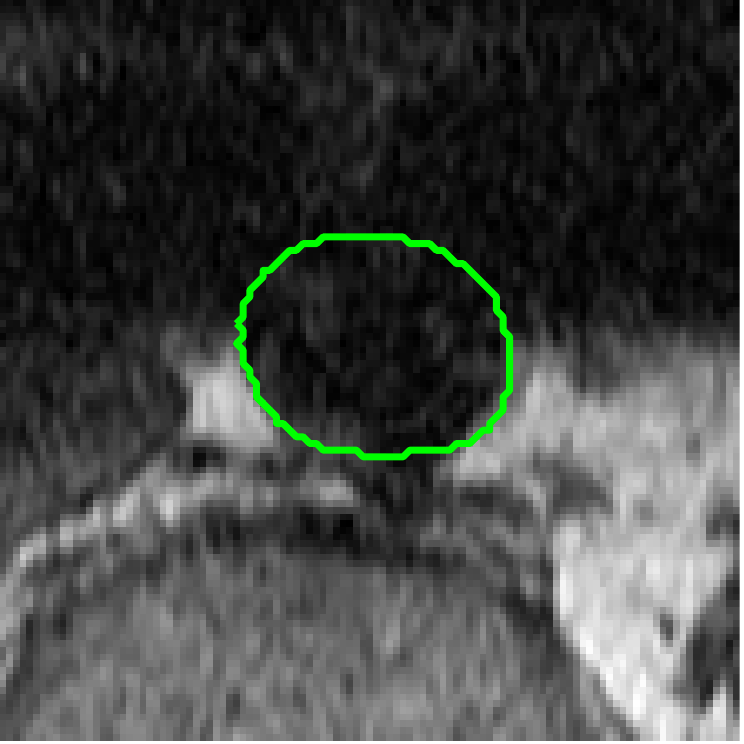}
        \hspace{-1.75 mm}
        \includegraphics[width=0.195\linewidth,height=0.195\linewidth]{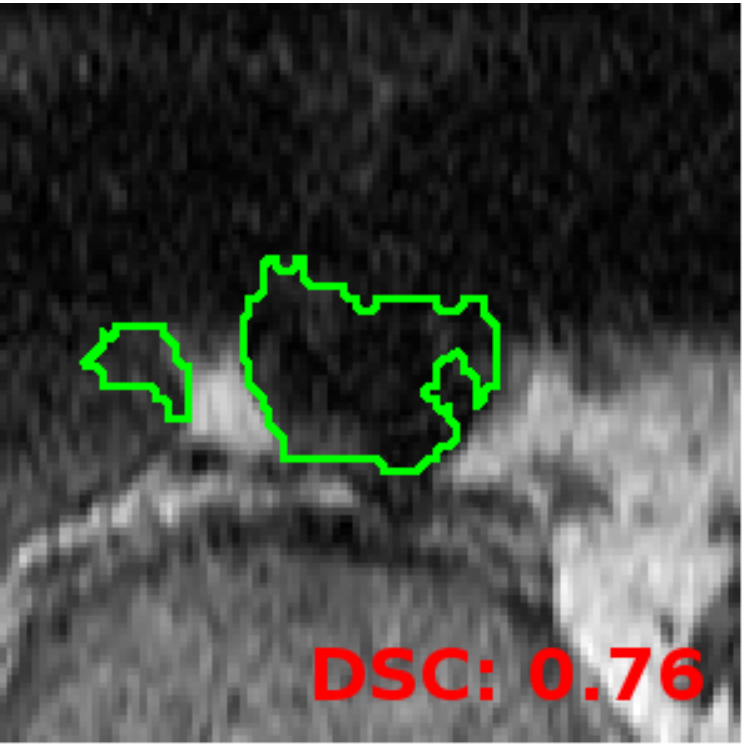}
        \hspace{-1.75 mm}
        \includegraphics[width=0.195\linewidth,height=0.195\linewidth]{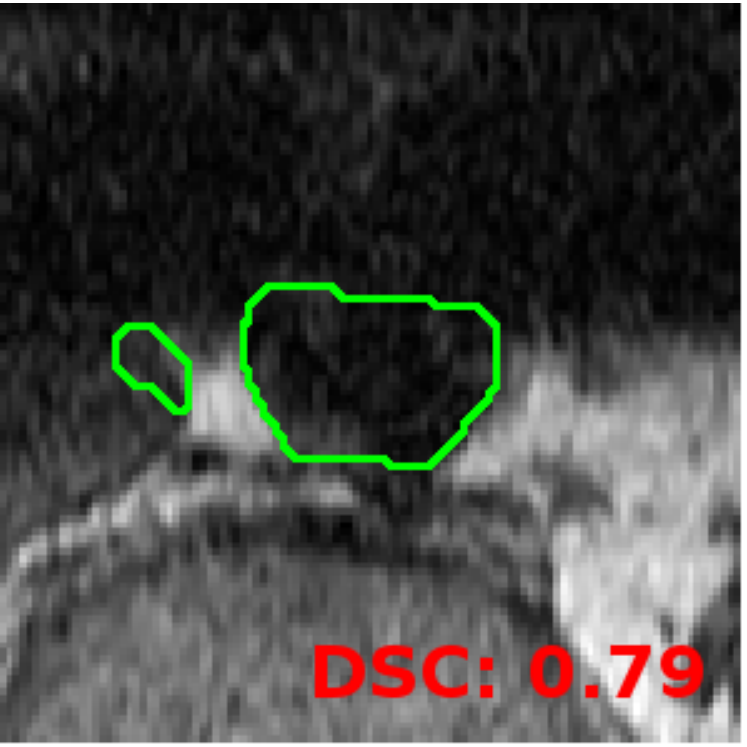}
        \hspace{-1.75 mm}
        \includegraphics[width=0.195\linewidth,height=0.195\linewidth]{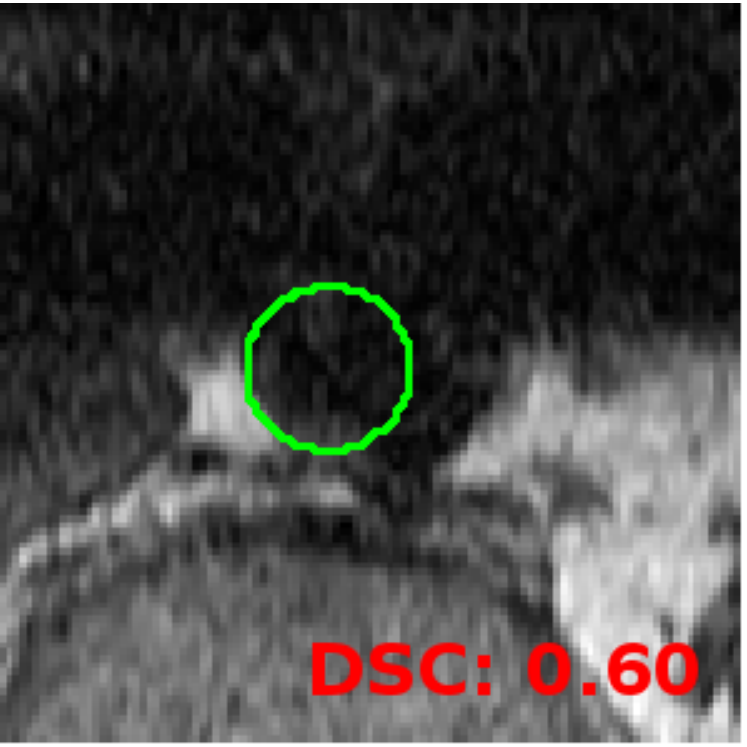}
        \hspace{-1.75 mm}
        \includegraphics[width=0.195\linewidth,height=0.195\linewidth]{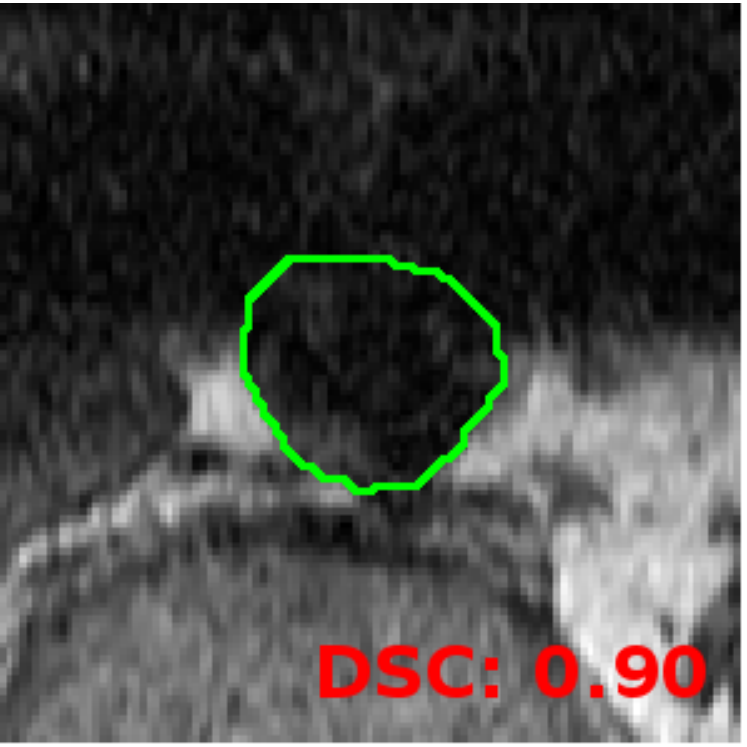}
        }

    
      \mbox{
        \includegraphics[width=0.195\linewidth,height=0.195\linewidth]{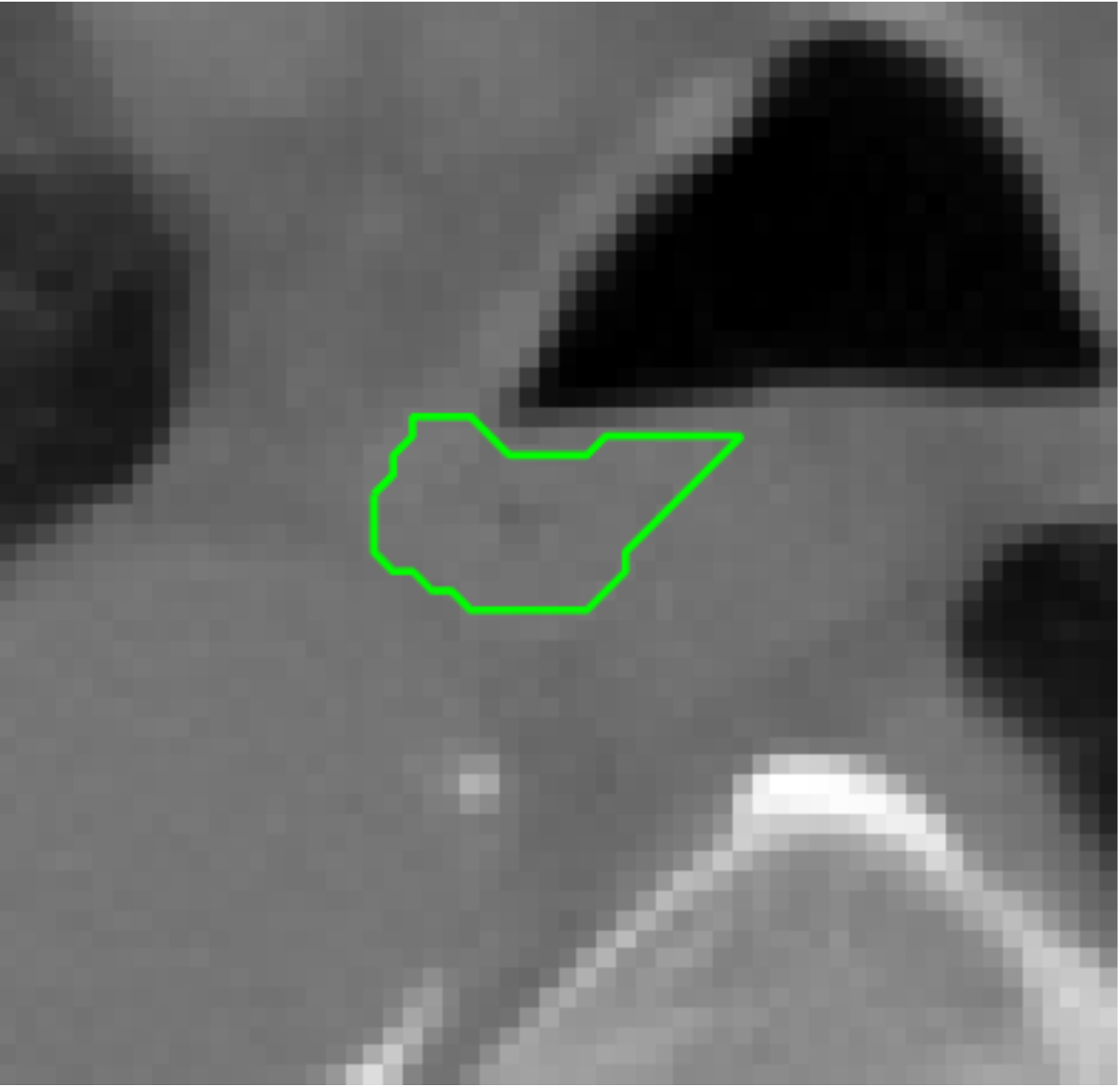}
        \hspace{-1.75 mm}
        \includegraphics[width=0.195\linewidth,height=0.195\linewidth]{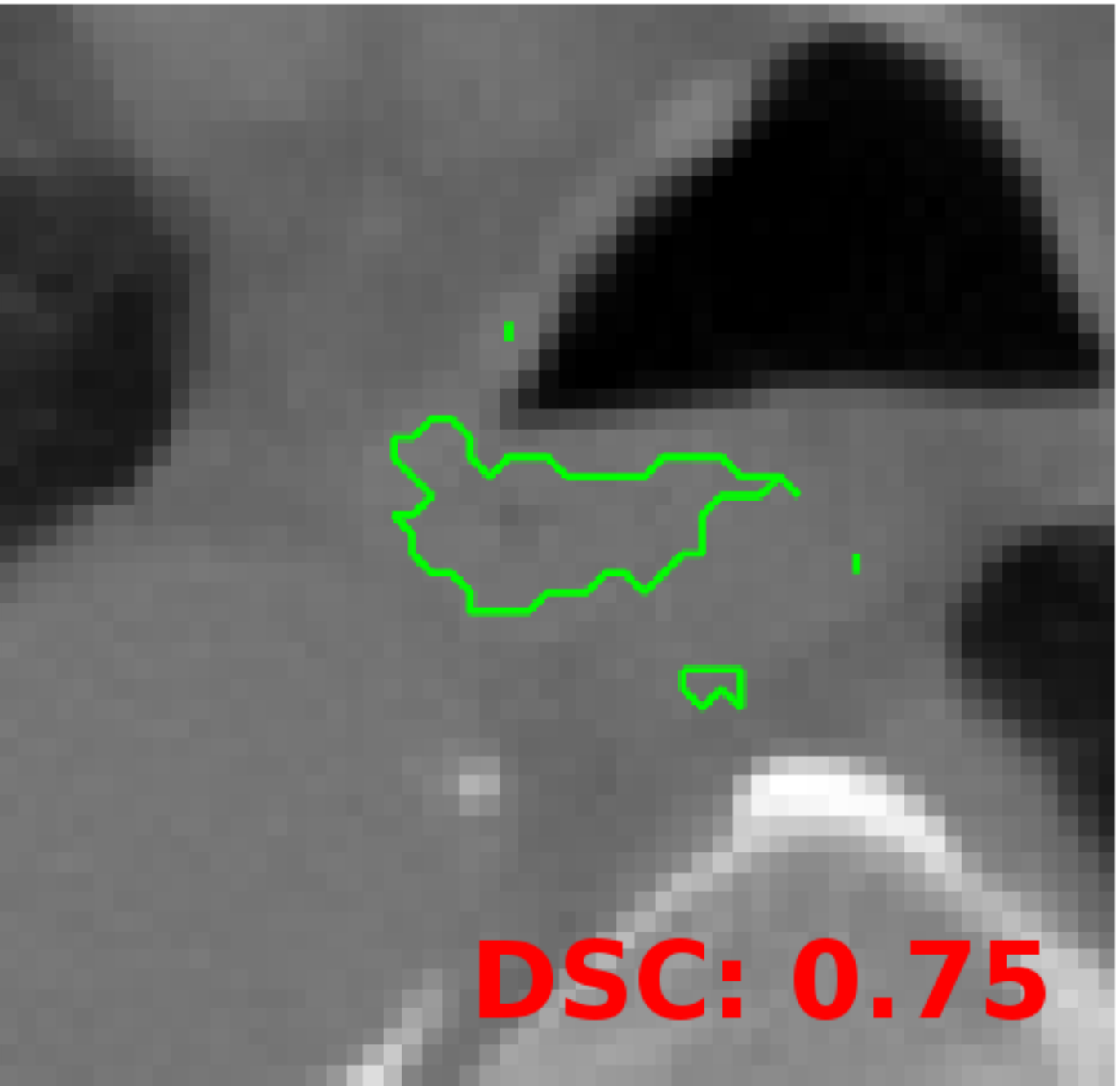} 
        \hspace{-1.75 mm}
        \includegraphics[width=0.195\linewidth,height=0.195\linewidth]{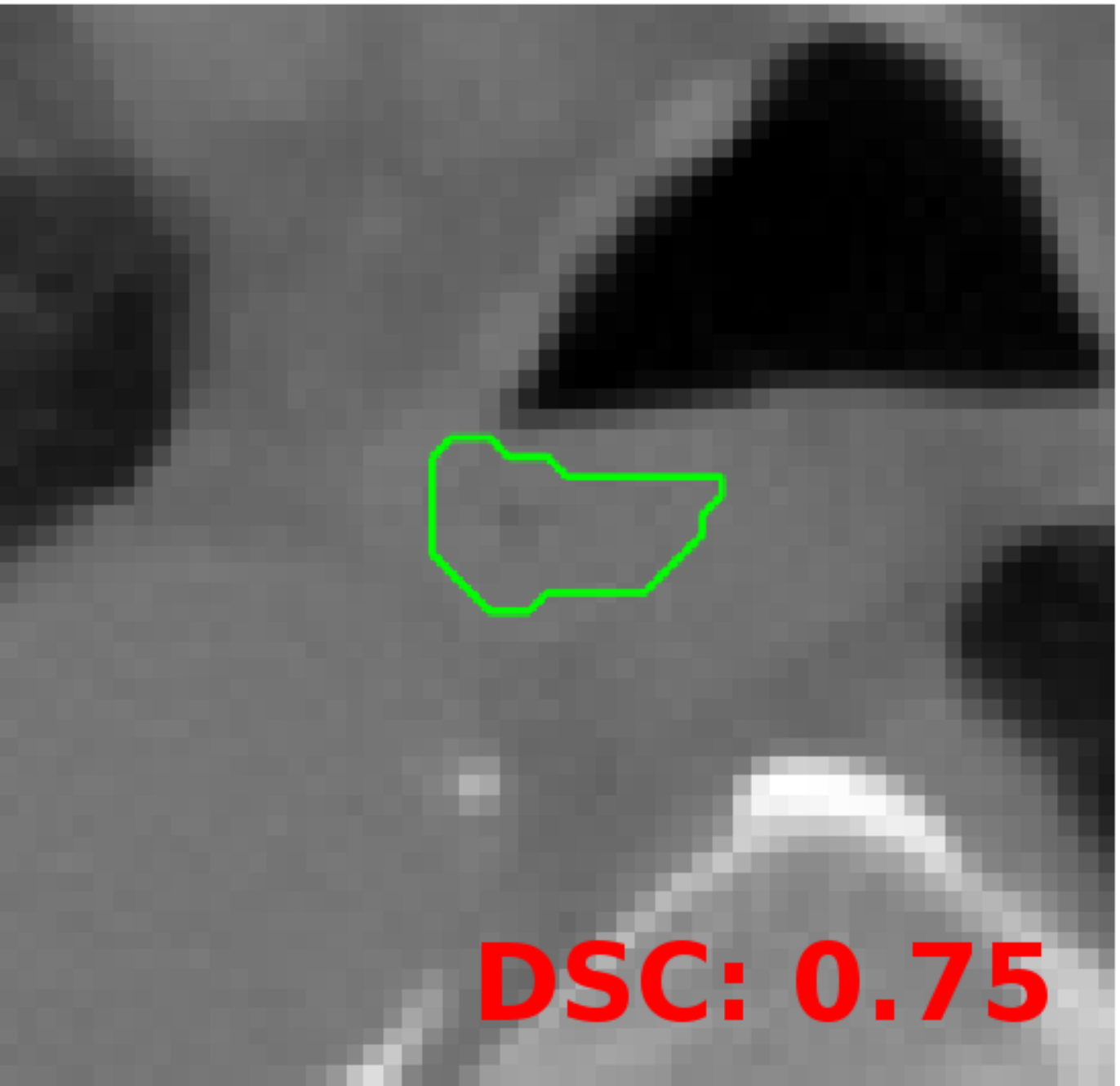}
        \hspace{-1.75 mm}
        \includegraphics[width=0.195\linewidth,height=0.195\linewidth]{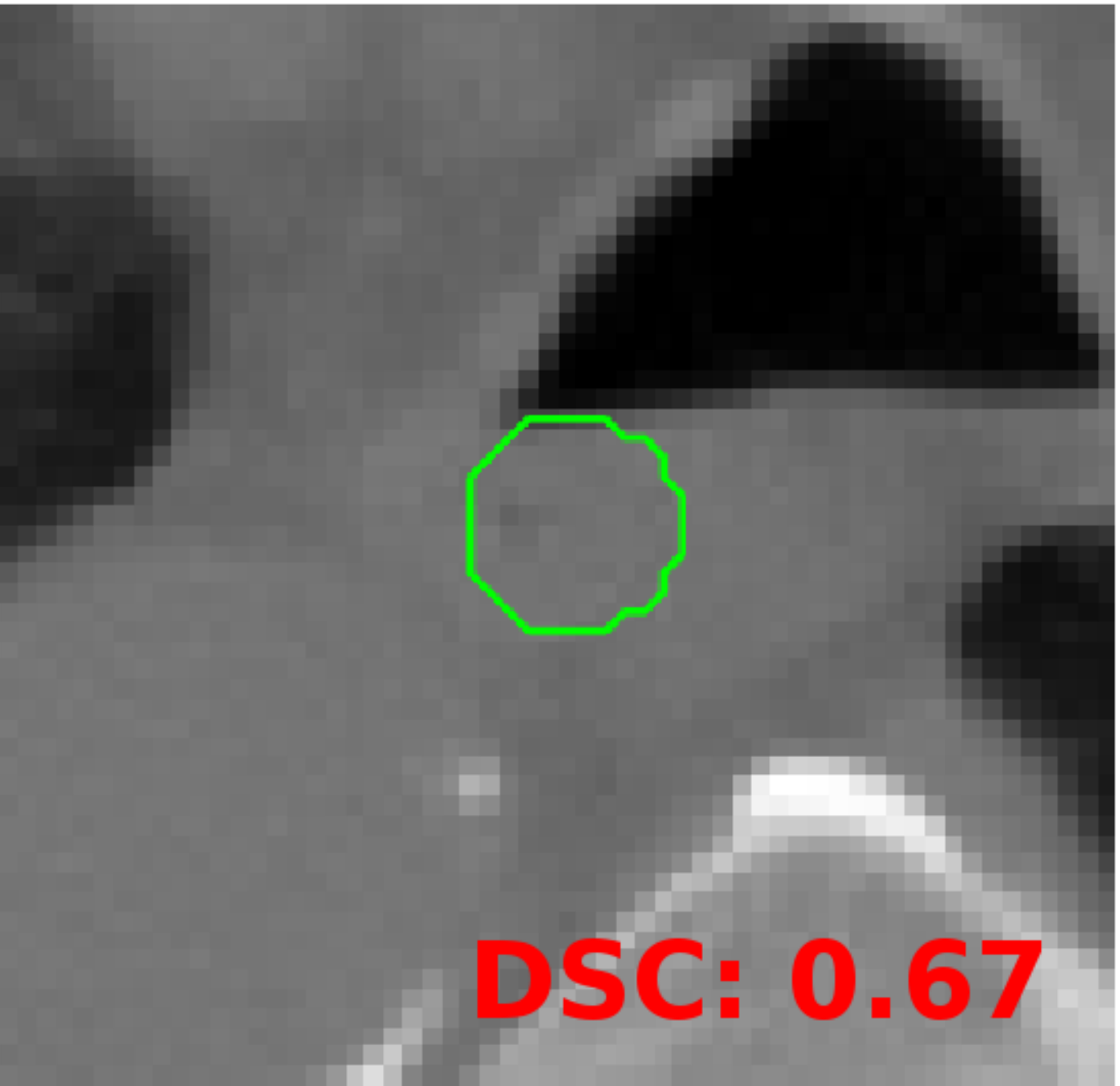} 
        \hspace{-1.75 mm}
        \includegraphics[width=0.195\linewidth,height=0.195\linewidth]{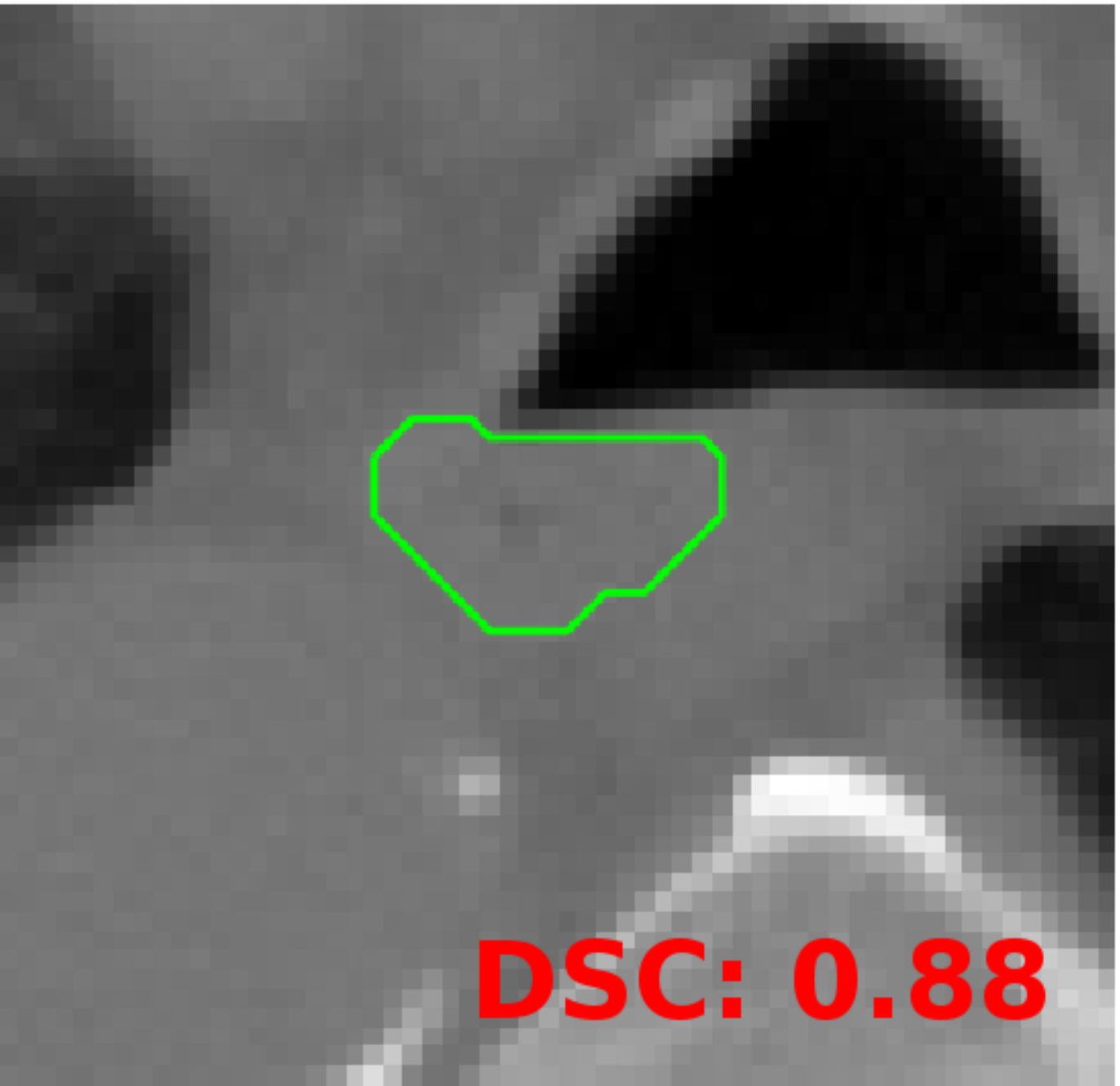}
        }
        
        \mbox{
        \includegraphics[width=0.195\linewidth,height=0.195\linewidth]{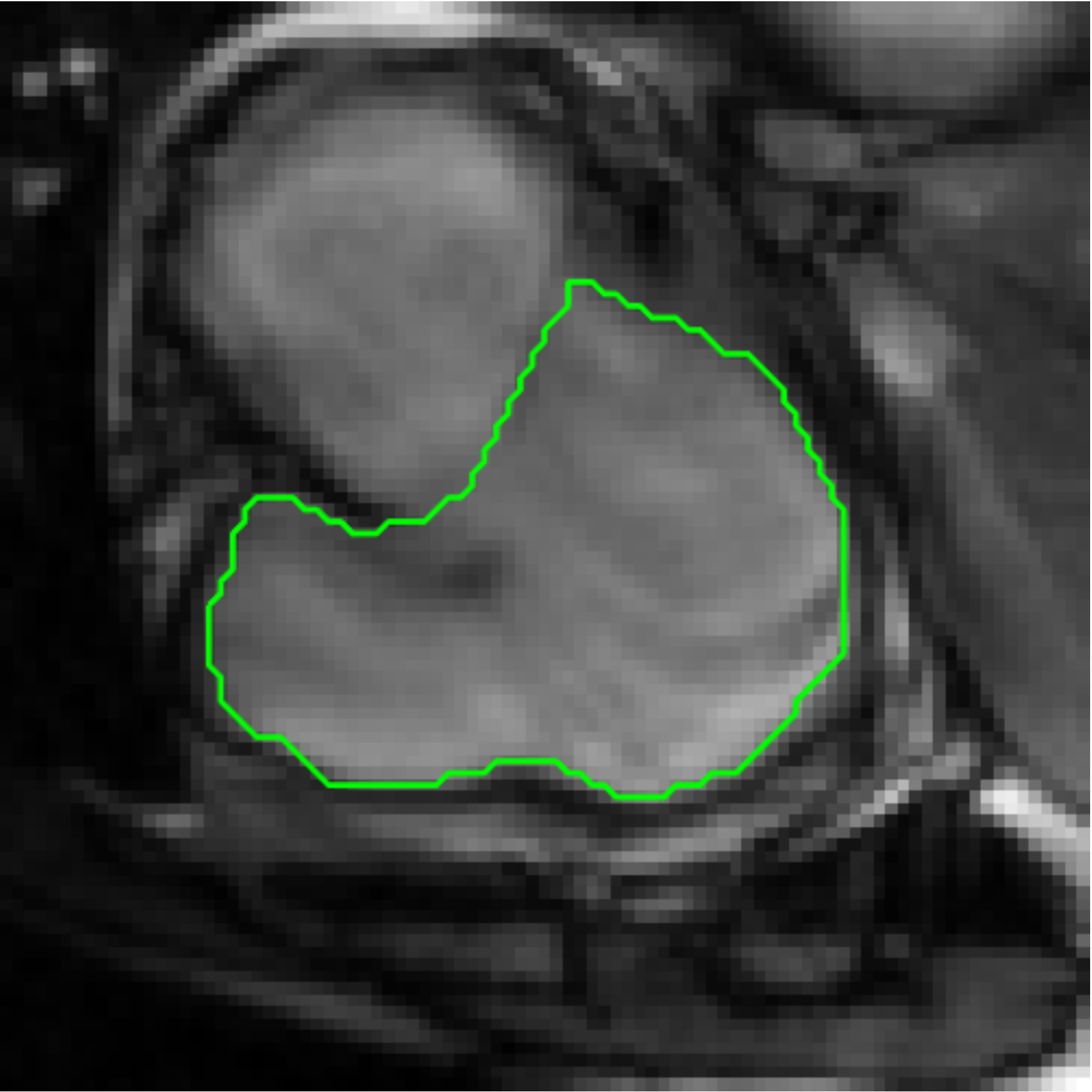}
        \hspace{-1.75 mm}
        \includegraphics[width=0.195\linewidth,height=0.195\linewidth]{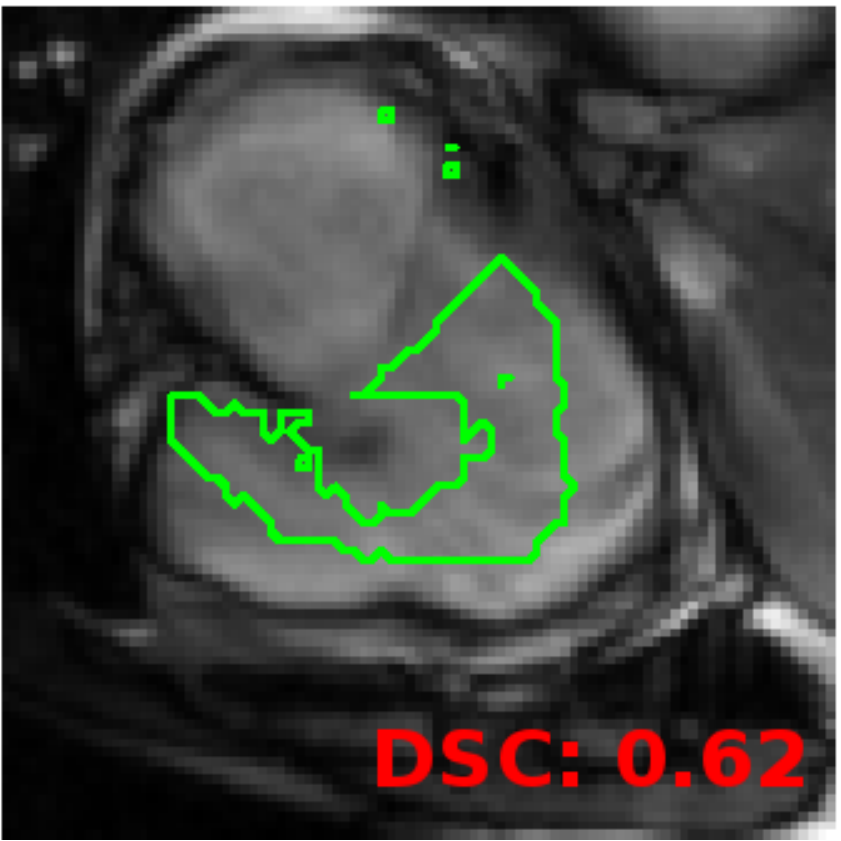}
        \hspace{-1.75 mm}
        \includegraphics[width=0.195\linewidth,height=0.195\linewidth]{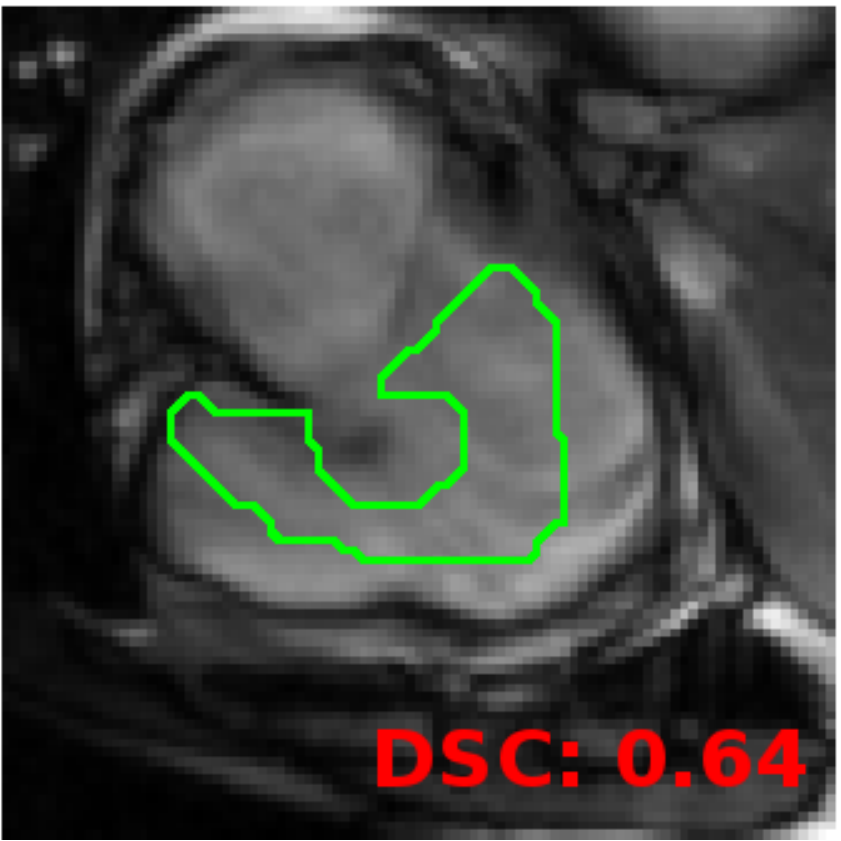}
        \hspace{-1.75 mm}
        \includegraphics[width=0.195\linewidth,height=0.195\linewidth]{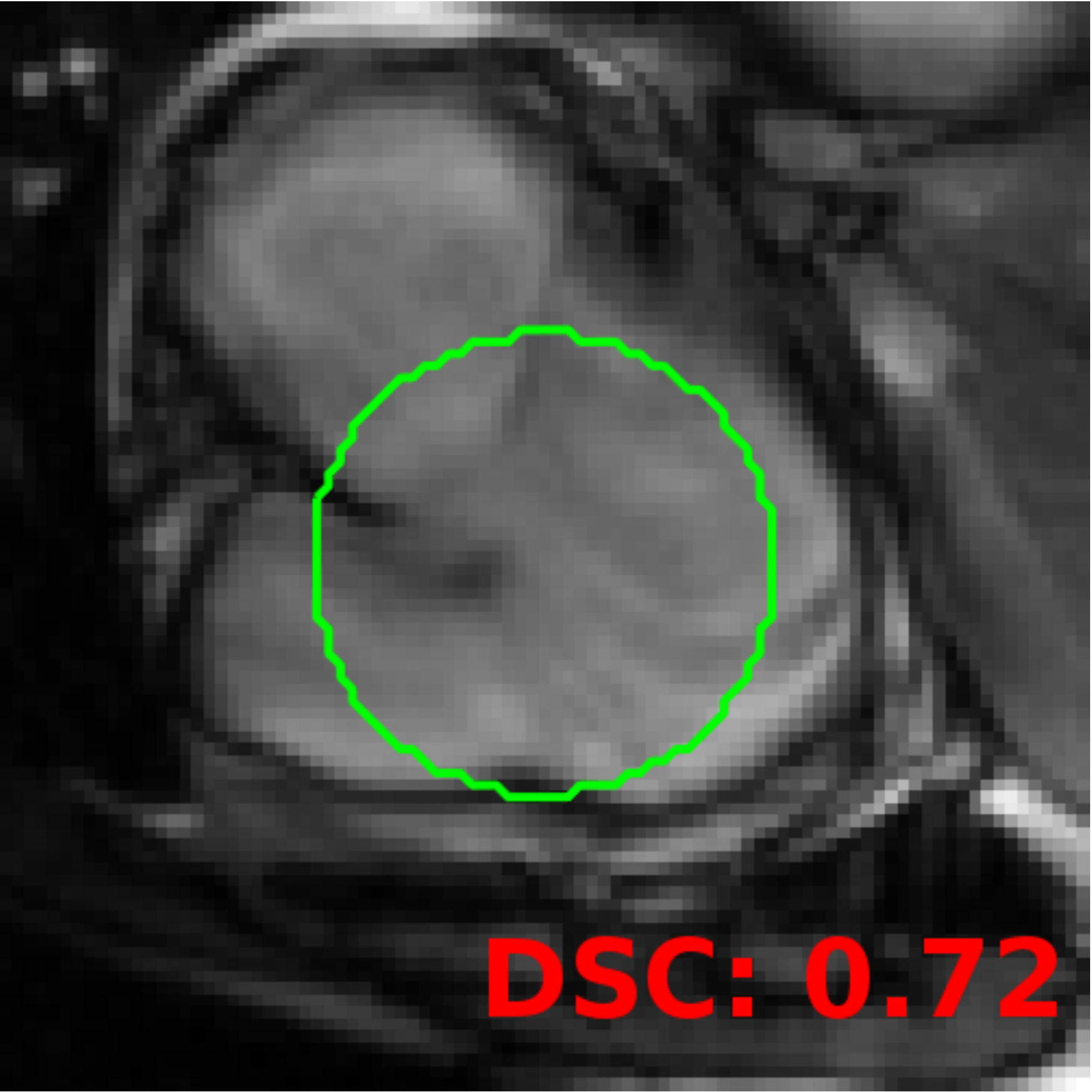}
        \hspace{-1.75 mm}
        \includegraphics[width=0.195\linewidth,height=0.195\linewidth]{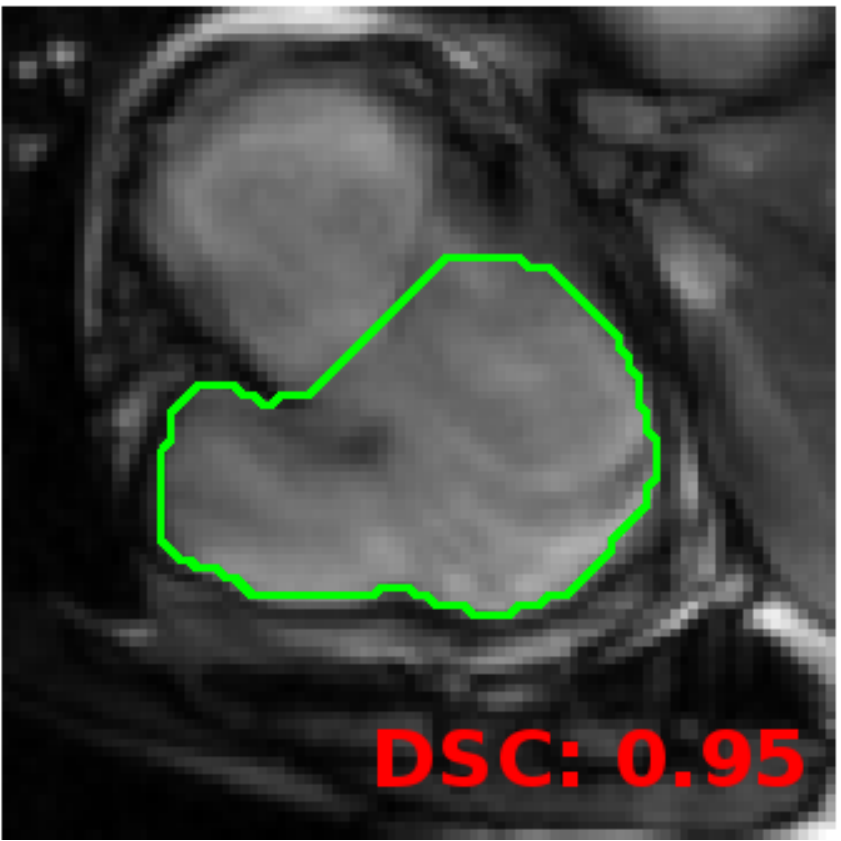}
        }

    \mbox{
      \shortstack{
        \includegraphics[width=0.195\linewidth,height=0.195\linewidth]{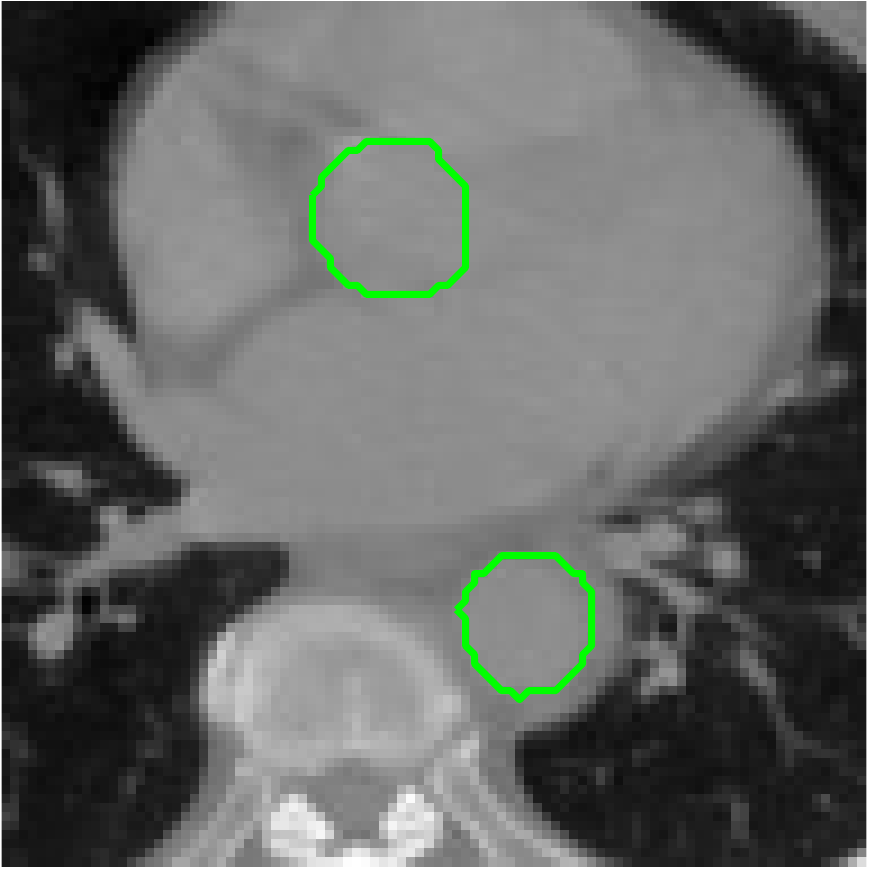} \\
        Ground truth 
        }
        \hspace{-1.75 mm}
      \shortstack{     
        \includegraphics[width=0.195\linewidth,height=0.195\linewidth]{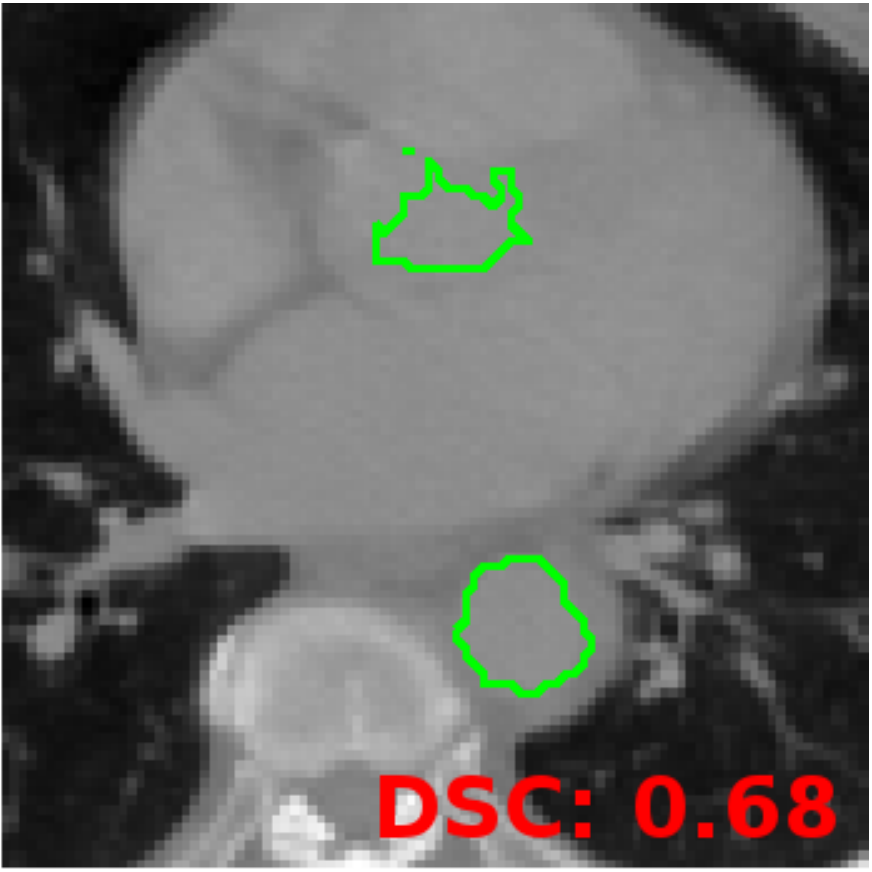} \\
        CNN \cite{dolz20163d} }
        \hspace{-1.75 mm}
      \shortstack{
        \includegraphics[width=0.195\linewidth,height=0.195\linewidth]{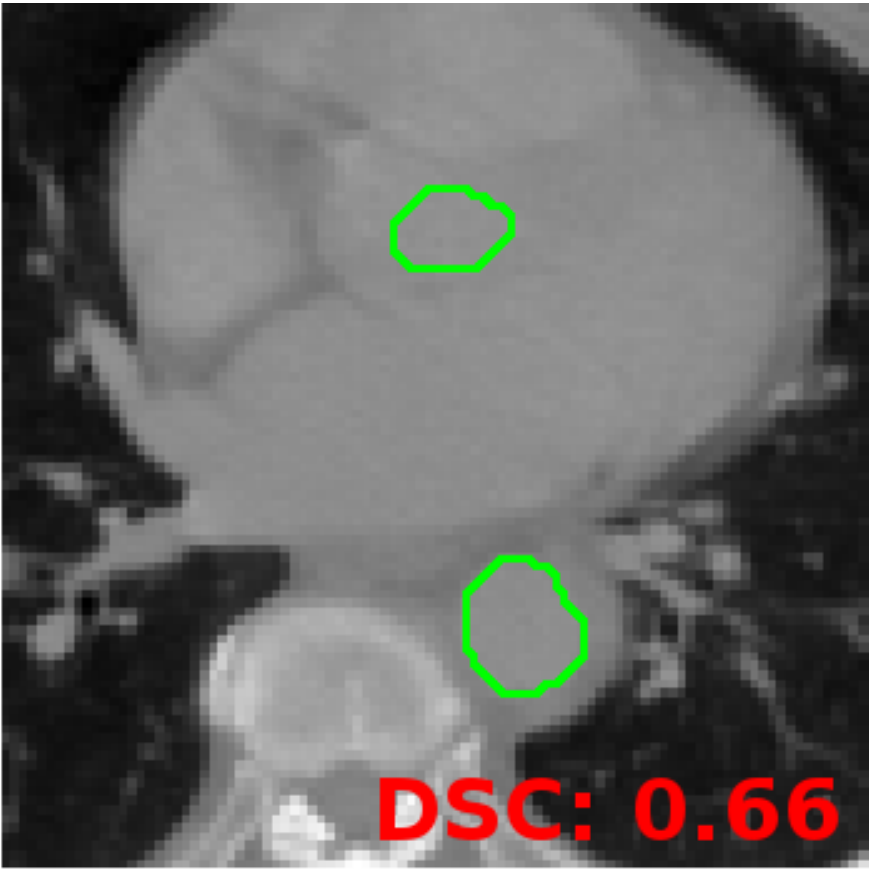} \\
        GCs \cite{Boykov2004}}
        \hspace{-1.75 mm}
         
      \shortstack{
        \includegraphics[width=0.195\linewidth,height=0.195\linewidth]{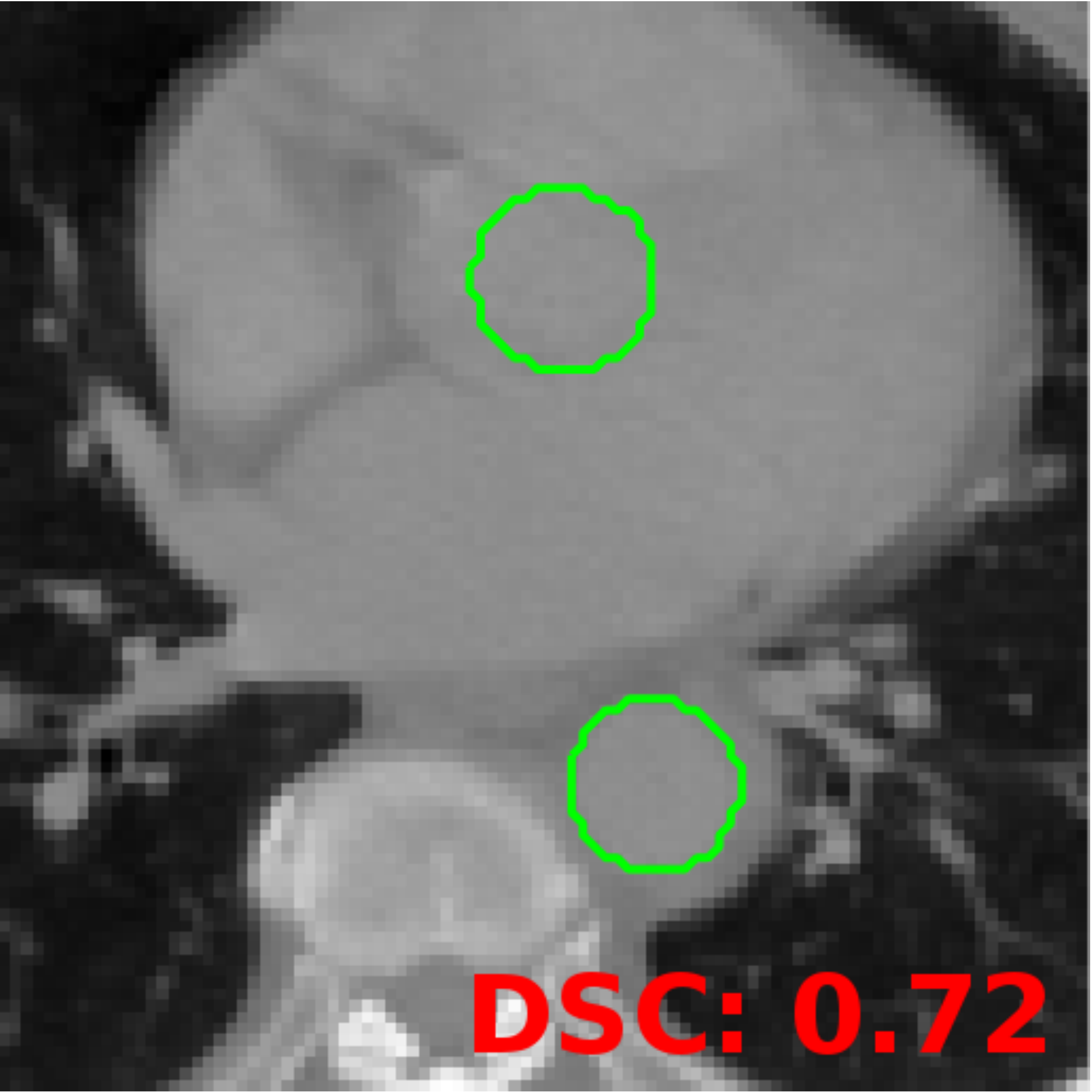} \\
        TRIC \cite{ayed2014tric}}
        \hspace{-1.75 mm}
         
      \shortstack{
        \includegraphics[width=0.195\linewidth,height=0.195\linewidth]{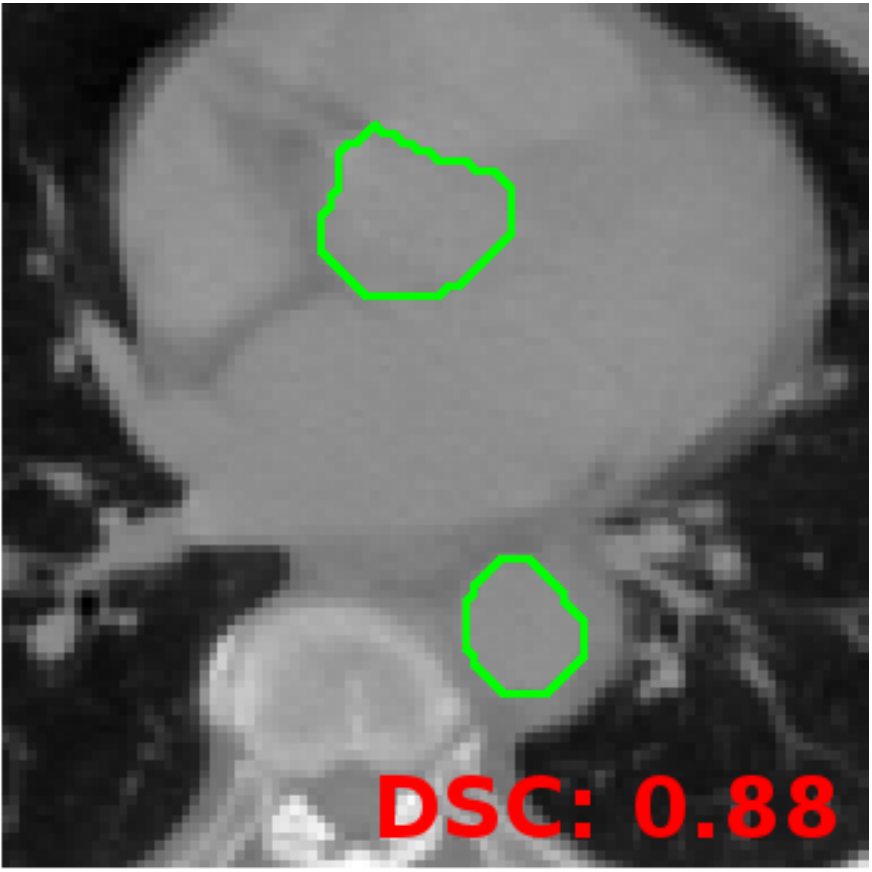} \\
        Proposed }
        
        }
        \caption{Segmentations results in several applications: aorta in MR-T1 (\emph{top}), esophagus in CT (\emph{second}), right ventricle in MRI (\emph{third row}) and aorta in CT.
        }
        \label{fig:ResultsMain}
\end{center}        
\end{figure}

Several recent studies have shown that generic shape constraints such as convexity \cite{Gorelick2017,veksler2008star}, compactness \cite{ayed2014tric,das2009semiautomatic}, axial symmetry \cite{qiu2013fast}, tubularity \cite{kitamura2014coronary}, skeleton consistency \cite{Isack2016} and inter-region topology \cite{BenTaied2016} can be very powerful in medical image segmentation. Such constraints can boost substantially the performances of state-of-the-art segmentation algorithms, including powerful supervised learning methods such as convolutional neural networks (CNN) \cite{BenTaied2016}. Imposing these constraints can be beneficial in a breadth of medical applications, particularly when training data is limited or when the target segments undergo strong noise and poor contrasts/resolutions (see Fig. \ref{fig:ResultsMain} for examples where target boundaries have varied shapes and low contrast). Unfortunately, these shape constraints are typically {\em high-order} 
functionals, which yield challenging optimization problems. 
For instance, the recent convexity term in \cite{Gorelick2017} involves a large number of non-submodular triple-cliques, and the compactness in \cite{ayed2014tric} is a high-order ratio, both requiring computationally expensive approximations and iterative schemes to reach a local minimum.

This study focuses on constraining segmentation functionals with shape compactness, a problem investigated previously in several works \cite{ayed2014tric,das2009semiautomatic,Kolmogorov2007,grady2006isoperimetric}.
In segmentation, the most common compactness functional is the ratio of boundary length to area (or surface to volume in 3D) \cite{Kolmogorov2007,grady2006isoperimetric}, which is related to the well-known {\em isoperimetric graph partitioning} problem (also referred to as the {\em Cheeger} problem). Unlike common length regularizers \cite{boykov2006graph}, this prior does not suffer from strong shrinking bias (i.e., bias towards small regions). Despite its simplicity, optimizing a discrete form of this ratio functional is an {\em NP-hard} problem \cite{Kolmogorov2007}. However, it is possible to transform this problem into a family of parameterized sub-modular pairwise costs, which can be explored efficiently with parametric max-flow optimizers \cite{Kolmogorov2007}. Unfortunately, length-to-area ratio varies significantly with region size\footnote{This is due to the fact that, in image segmentation, region size might be several orders of magnitude larger then boundary length.}. As pointed out (and observed experimentally) in several studies \cite{grady2006isoperimetric,Kolmogorov2007}, the isoperimetric partitioning problem has a strong bias towards large regions, which is a serious limitation in practice. 

The recent segmentation study in \cite{ayed2014tric} proposed an alternative compactness measure based on a ratio of regional moment statistics, thereby ensuring size (or scale) invariance. This algorithm showed promising performances in the case of a vascular structure, and can handle arbitrary variations in region size. Unfortunately, this compactness term is pose-dependent, requiring the user to specify the center of mass of the target region in each 2D slice of a subject data set. This precludes its use for fully automated segmentation, particularly in the case of large 3D volumes. More importantly, it is strongly biased towards circular shapes, significantly limiting its applicability (see the examples in Fig. \ref{fig:ResultsMain}). Another related work is the semi-automatic segmentation algorithm in \cite{das2009semiautomatic}, whose compactness prior also favors large regions and requires user seeds.

The contributions of this study can be summarized as follows.
\begin{itemize} 
\item {\bf General-purpose segmentation:} We propose to constrain segmentation functionals with a dimensionless, unbiased and position-independent shape compactness prior, which we solve efficiently with an alternating direction method of multipliers (ADMM). To the best of our knowledge, the ratio of {\em length-squared} to area has not been used before in segmentation functionals, although it is widely accepted as a compactness measure in the area of shape metrics \cite{Santiago2009}. This is probably due to the fact that the ensuing high-order optimization problem is challenging. Involving a squared sum of pairwise potentials, our problem makes the use of powerful sub-modular optimization techniques (e.g., parametric max-flow) impractical for image segmentation problems, as it would result in dense (fully connected) graphs. 
We propose an efficient alternative by splitting the problem into a sequence of easier sub-problems. At each iteration, our ADMM solution alternates three steps, each can be performed efficiently: (i) a sparse-matrix inversion based on Woodbury identity, (ii) a closed-form solution of a cubic equation and (iii) a graph-cut update of a sub-modular pairwise sub-problem involving a sparse graph.  

\item {\bf Clinical applications:} We first report a comprehensive evaluation of our algorithm on the challenging task of abdominal aorta segmentation in 3D MRI. Results obtained on a dataset of 40 subjects demonstrate a state-of-the-art performance of the proposed method, with an average Dice metric of 0.81. Except for the recent semi-automatic algorithms described in \cite{duquette20123d,ayed2014tric}, most existing aorta segmentation techniques have focused on CTA. Unlike these techniques, our algorithm allows the development of fully automated tools, which could be used retrospectively on large-scale MRI datasets to identify potentially fatal abnormalities like abdominal aortic aneurysms (AAA) \cite{Georgakarakos2011}. Our experiments further show the usefulness of our method in three important and challenging segmentation problems: the delineation of the esophagus in CT, the right ventricle in MR, and the aorta in CT.

\end{itemize}

\section{Formulation}



Let $\Omega$ denote the image domain and $\xx_i \in \RR^K$ the input-feature vector of pixel (or voxel) $i \in \Omega$. 
Segmentation assigns to each $i$ a label $y_i \in \LL$ ($\LL$ is the set of possible labels). While our formulation could be easily extended to multi-class segmentation, we focus on a binary (two-region) statement for a clearer presentation. 
We minimize the following functional with respect to a discrete binary vector $\yy \in \{0,1\}^{|\Omega|}$:
\beq \label{eq:energies}
    E(\yy) \, = \, E_p(\yy) \, + \, \lambda \, E_c(\yy), 
\eeq
where $E_p$ encodes unary-potential priors, and $E_c$ is our shape compactness.
As is typical in segmentation, $E_p$ can be expressed in term of log-likelihoods: $E_p(\yy) \, = \, \sum_i u_i \, y_i$,  
where $u_i = \log p(y_i=0 \, | \, \xx_i) - \log p(y_i=1 \, | \, \xx_i)$. Although any (semi-) supervised technique can be used to learn $p(y_i \, | \, \xx_i)$, in this study, we employed a fully-convolutional neural network (FCNN) \cite{dolz20163d}.

Our dimensionless, unbiased and position-independent shape compactness prior is based on a measure that is well accepted 
in the context of shape metrics \cite{Santiago2009}, the ratio of {\em length-squared} to area, 
which we can write in discrete form (i.e., as a function of our segmentation variable): 
$E_c(\yy) = P(\yy)^2 / A(\yy)$. The area can be expressed as $A(\yy) = \sum_i y_i$, and the length is proportional to the number of neighboring pixels with different labels, i.e., $P(\yy) \propto \sum_{i,j} w_{ij} (y_i - y_j)^2$,
where pairwise potential $w_{ij}=1$ if $i$ and $j$ are neighbors, otherwise $w_{ij}=0$. Note that this compactness measure can be easily extended so as to attract the solution towards strong edges in the image: $w_{ij} = \exp(-\sum_k \sigma_k \, (x_{ik} - x_{jk})^2)$, with
$\sigma_k$ controlling the relative importance of feature $k$ on the weight.
With these definitions of $E_p$ and $E_c$, and re-writing length as $P(\yy) = \tr{\yy}L\yy$, where $L$ is the Laplacian matrix corresponding to weights $w_{ij}$, our compactness model becomes:
\beq\label{eq:comp_problem}
	\argmin_{\yy \, \in \, \{0,1\}^{|\Omega|}} \ 
	    E(\yy) \ = \ \tr{\uu}\yy \, + \, \lambda \frac{\big(\tr{\yy}L\yy \big)^2}{\tr{\vone}\yy},
\eeq 
where $\vone$ is a vector with the value one for each element. 

\subsection{ADMM optimization} 

The general principle of ADMM is to decompose a hard problem into easier-to-solve sub-problems, which are coupled together via equality constraints. In our case, we introduce auxiliary variable $\zz \in \RR^{|\Omega|}$ and $\cc \in \RR_+$, and reformulate problem (\ref{eq:comp_problem}) as  
\beq\label{eq:ADMM1}
    \argmin_{\yy, \, \zz, \, \cc} \ 
        \tr{\uu}\yy \, + \, \frac{\lambda}{\cc} \big(\tr{\yy}L\yy\big)\big(\tr{\zz}L\zz\big), 
            \quad \tx{s.t.} \  \yy = \zz \, \tx{ and } \, \cc = \tr{\vone}\zz.
\eeq
Note that since we impose $\yy = \zz$, we can relax the binarity constraints on $\zz$. This strategy allows an efficient update of this variable.

In the next step, we move the constraints in the cost-function via a augmented Lagrange formulation:
\beq \label{eq:ADMM2}
    \argmin_{\substack{\yy, \, \zz, \, \cc \\ \vnu_1, \, \nu_2}} \ 
        \tr{\uu}\yy \, + \, \frac{\lambda}{\cc} \big(\tr{\yy}L\yy\big)\big(\tr{\zz}L\zz\big) 
            \, + \, \frac{\mu_1}{2}\big\|\yy - \zz + \vnu_1\big\|_2^2 
                \, + \, \frac{\mu_2}{2}\big(\cc - \tr{\vone}\zz + \nu_2\big)^2.
\eeq
In this new formulation, $\vnu_1$ and $\nu_2$ are dual variables corresponding to the two constraints, and $\mu_1$, $\mu_2$ parameters controlling the trade-off between the primal objective and constraint satisfaction. The benefit of this formulation is that solving for each variable, considering all other variables fixed, can be carried out efficiently and to optimality. We thus adopt an iterative optimization method, where each variable is updated in turn until convergence. 

\paragraph{Updating $\zz$:} Let $\alpha = \frac{\lambda}{\cc}\,\tr{\yy}L\yy$, the task of updating $\zz$ can be expressed from (\ref{eq:ADMM2}) as
\beq 
    \argmin_{\zz \, \in \, \RR^{|\Omega|}} \ \alpha\tr{\zz}L\zz 
        \, + \, \frac{\mu_1}{2}\big\|\zz-(\yy+\vnu_1)\big\|_2^2 
            \, + \, \frac{\mu_2}{2}\big(\tr{\vone}\zz - (\cc+\nu_2)\big)^2.
\eeq
Minimizing this convex quadratic problem yields
\beq
    \zz \ = \ \big(\alpha L + \mu_{1} I + \mu_2 \vone\tr{\vone}\big)^{-1}  
        \big( \mu_1(\yy+\vnu_1) \ + \ \mu_2(\cc+\nu_2) \vone \big).
\eeq
Note that computing $\zz$ requires solving a large and dense linear system. Let $Q = \alpha L + \mu_1 I$, following the Woodbury identity and using the fact that $Q^{-1}\vone = \frac{1}{\mu_1}\vone$, we can reformulate the matrix inversion as:
\begin{align}
    \big(\alpha L + \mu_1 I + \mu_2 \vone\tr{\vone}\big)^{-1} & \, = \,
        Q^{-1} \, - \, {\Big(\frac{1}{\mu_2} + \tr{\vone} Q^{-1} \vone\Big)}^{-1} 
            Q^{-1} \vone \tr{\vone}Q^{-1}\nonumber\\
    & \, = \, Q^{-1} \, - \, \frac{1}{\mu_1}{\Big(\frac{\mu_1}{\mu_2} + |\Omega|\Big)}^{-1} \vone\tr{\vone}. 
\end{align}
Since $Q$ is very sparse, the resulting system can be solved efficiently via standard techniques like the preconditioned conjugate gradients method.

\paragraph{Updating $\cc$:} Let $\beta = \lambda (\tr{\yy}L\yy) (\tr{\zz}L\zz)$, updating $\cc$ amount to solving
\beq 
    \argmin_{\cc \geq 0} \ \frac{\beta}{\cc} \, + \, \frac{\mu_2}{2} \big(\cc - (\tr{\vone}\zz - \nu_2)\big)^2
\eeq
Deriving this equation w.r.t. $\cc$ and setting the result to zero gives the following cubic equation:
\beq 
   \cc^3 \, - \, \big(\tr{\vone}\zz - \nu_2\big)\cc^2 \, = \, \frac{\beta}{\mu_2} \, > \, 0. 
\eeq
It can be shown that this equation has a real valued root $\cc > \tr{\vone}\zz - \nu_2$. In practice, $\tr{\vone}\zz - \nu_2$ is positive and converges toward zero, and $\cc$ will be positive. 

\paragraph{Updating $\yy$:} Let $\gamma = \frac{\lambda}{\cc}\tr{\zz}L\zz$ and $\qq = \zz-\vnu_1$. We update $\yy$ by considering the following problem:
\beq \label{eq:updateY1}
    \argmin_{\yy \, \in \, \{0,1\}^{|\Omega|}} \ 
        \tr{\uu}\yy \, + \, \lambda \tr{\yy}L\yy \ + \ \frac{\mu_1}{2} \big\|\yy - \qq \big\|_2^2
\eeq
Since $\yy$ is binary, we have that $y_i^2 = y_i$ and $(y_i - y_j)^2 = |y_i - y_j|$. Hence, we can reformulate (\ref{eq:updateY1}) as
\beq
\argmin_{\yy \, \in \, \{0,1\}^{|\Omega|}} \ 
    \sum_{i \in \Omega} \big(u_i + \mu_1(\tfrac{1}{2} - q_i)\big) y_i \, + \, 
        \lambda \sum_{i,j \in \Omega^2} w_{ij} \big|y_i - y_j\big|.
\eeq
This corresponds to a simple graph-cut problem, which can be solved efficiently using the Boykov-Kolmogorov algorithm \cite{Boykov2004}.

Finally, we update dual variables following the standard ADMM algorithm: $\vnu'_1 =  \vnu_1 + (\yy-\zz)$ and $\nu'_2 = \nu_2 + (\cc - \tr{\vone}\zz)$. This iterative updating process is repeated until $\|\yy - \zz\|_2$ is less than a small epsilon.




\vspace{-10mm}
\section{Experiments}
\vspace{-1mm}

We first present a quantitative evaluation of our shape compactness on the task of segmenting the abdominal aorta in a MR-T1 data set of 40 subjects. The usefulness of the proposed method is then shown qualitatively for three additional segmentation scenarios: CT esophagus, MRI right ventricle and CT aorta segmentation. Our method's accuracy is measured in terms of Dice coefficient, and compared to that of graph cuts (GC) \cite{Boykov2004} and TRIC\footnote{Note that TRIC needs a reference point to define the shortest-path distance. This point was defined as the centroid of the CNN-based segmentation in this work.} compactness \cite{ayed2014tric}. To measure the contribution of the pixelwise (i.e., unary potential) prior, we also report results obtained via a simple thresholding.

While any technique can be used to obtain the unary-potentials in Eq. (\ref{eq:comp_problem}), we considered the 3D fully-CNN (FCNN) architecture presented in \cite{dolz20163d}. This architecture is comprised of 9 convolutional layers with non-linear activation units, 3 fully-connected layers (converted into standard convolution operations), and a soft-max layer. The output of this model is a map indicating the probability of each pixel to belong to a given structure. The same unary potential was used for all three tested approaches. For all segmentation applications, the FCNN was trained using a \textit{k}-fold validation strategy, with $k=4$.

For our method's ADMM parameters, we used $\mu_{1}\!\approx\!2000$ and $\mu_{2}$=50 for all experiments. As mentioned in \cite{boyd2011distributed}, ADMM algorithms are not overly sensitive to these parameters. To facilitate convergence, we increase the value of these parameters by 1\% (i.e., 1.01 multiplication factor) at each iteration. In contrast, the compactness regularization $\lambda$ was tuned for each segmentation problem: $\lambda$ = 5000 for MR-T1 abdominal aorta, $\lambda$ = 1000 for CT esophagus, $\lambda$ = 15000 for MR right ventricle, and $\lambda$ = 3000 for CT abdominal aorta. The code has been made publicly available at \href{https://github.com/josedolz/UnbiasedShapeCompactness}{ https://github.com/josedolz/UnbiasedShapeCompactness}.

\vspace{-5mm}

\begin{figure}[ht!]
    \begin{center}
        \includegraphics[width=\linewidth]{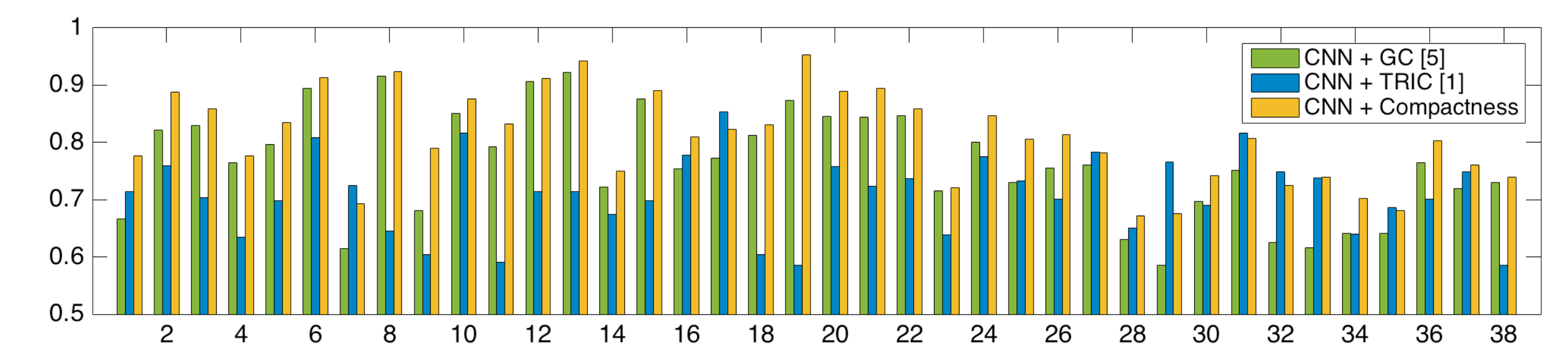} \\

\caption{Dice metric for 38 subjects of the 3D FCNN \cite{dolz20163d} output with graph cut\cite{Boykov2004} regularization, with TRIC compactness \cite{ayed2014tric}, and with the proposed compactness. }
\label{fig:dsc}
\end{center}        
\end{figure}

\vspace{-5mm}

\textbf{MR-T1 abdominal aorta segmentation:} As shown in the first row of Fig. \ref{fig:ResultsMain}, segmenting the aorta from MR-T1 images is a very challenging problem due to the noise and lack of visible boundaries. In this example, thresholding the FCNN probabilities yields parts of the background along with irregular contours. Because it regularizes the segmentation, GC provides a smoother contour. However, this contour follows the probability map of FCNN and does not reflect the compact shape of the target region. Increasing regularization weight in GC yields a compact region but decreases performance significantly due to the well-known shrinkage bias of standard pairwise length regularization (bias to small regions); note that regularization weight was tuned for an optimal GC performance. The next experimental examples will further highlight shrinkage bias. Unlike GC, TRIC finds a more compact region, due to its shape circularity prior. Nonetheless, this example illustrates this approach's two main drawbacks: 1) it is biased to near-circular regions, and 2) it is not translation invariant and finds regions centered on the probability map's center of mass. As seen in the figure, our pose-independent method did not suffer from a circularity/shrinkage bias, yielding a much more accurate segmentation, highly similar to the reference contours. 

The bar plot of Fig. \ref{fig:dsc} gives the accuracy of tested methods obtained for each subject\footnote{Two subjects were excluded from the quantitative analysis, as the FCNN failed to generate a usable probability map.}. The mean Dice and standard deviation computed across all subjects, is reported in Table \ref{tab:res}. These results confirm that the proposed compactness term provides a considerable improvement in accuracy. In terms of runtime, our method takes about 80 seconds on average to segment a single subject, 60 seconds of time spend computing the unary potentials via the FCNN. In contrast, TRIC requires nearly 20 minutes per subject. Hence, our method is more suitable for the fully automated segmentation of large-scale datasets.


\vspace{-5mm}

\begin{table}[ht!]
\centering
\begin{small}
\begin{tabular}{C{25mm}C{20mm}C{25mm}C{20mm}C{20mm}}
\toprule
\textbf{Method}  & \textbf{CNN \cite{dolz20163d}} & \textbf{CNN + GC \cite{Boykov2004}} & \textbf{TRIC \cite{ayed2014tric}} & \textbf{Proposed} \\ 
\midrule\midrule
\textbf{Dice} & 0.73 (0.09)   & 0.76 (0.09) & 0.71 (0.07) & 0.81 (0.08)     \\
\textbf{Time} & $\sim$ 60 sec & $\sim$ 70 sec & $\sim$ 20 min & $\sim$ 80 sec \\ 
\bottomrule
\end{tabular}
\end{small}
\caption{Quantitative evaluations of MR-T1 abdominal aorta segmentations.}
\label{tab:res}
\end{table}

\vspace{-10mm}

\textbf{CT esophagus segmentation:} Fig. \ref{fig:ResultsMain} (\textit{second row}) shows delineations of the esophagus in CT, also a challenging problem due to its complex shape and non-homogeneous appearance. In this particular example, we can observe how the well-known shrinkage bias problem of GC leads to an under-segmentation of the esophagus. Notice that GC yielded a compact region but did not improve FCNN's performance. As in the previous case, we also observe how TRIC over-enforces shape circularity on the segmentation. On the other hand, our compactness term is able to preserve the target region's shape. 



%

%
%

\textbf{MR Right Ventricle Segmentation:} This qualitative result (Fig. \ref{fig:ResultsMain}, \textit{third row}) shows how our compactness term can accommodate a more general class of shapes that differs significantly from tubular structures. Although TRIC \cite{ayed2014tric} can also handle shapes different than a circle, multiple reference points, which form a skeleton, are required from the user in this case, a prohibitively time-consuming effort for 3D data. However, to keep the process fully automatic, only the centroid of the FCNN segmentation was provided. Having only a single reference point, and particularly in this scenario, led to failure of TRIC to achieve a satisfactory segmentation.

         

%
        

%
%
%

\textbf{CT Aorta Segmentation:} The application of the proposed method to CT aorta segmentation is illustrated in the last row of Fig \ref{fig:ResultsMain}. We see that our method can handle multi-region scenarios, i.e. bifurcations, in contrast to other compactness terms \cite{ayed2014tric}. Note that, for this example, we applied TRIC separately on each of the two regions to avoid having a single contour centered in between these regions. This means that we fed TRIC with additional supervision information, unlike our method which is fully automatic. Notice that, to obtain compact and smooth regions, GC worsened FCNN performance due to shrinkage bias.

\section{Conclusion}

We presented an unbiased, fully-invariant and multi-region prior for the segmentation of compact shapes, based on the ratio of length-squared to area. An efficient ADMM strategy was proposed to solve the high-order energy minimization problem resulting from this formulation. Using an FCNN to obtain unary probabilities, the proposed method achieved high accuracy in four challenging segmentation problems.


\bibliographystyle{splncs03}
%

\end{document}